\pdfoutput=1
\documentclass[11pt]{article}

\usepackage{acl}

\usepackage{times}
\usepackage{latexsym}

\usepackage[T1]{fontenc}
\usepackage[utf8]{inputenc}

\usepackage{microtype}

\usepackage{graphicx}
\usepackage{amsmath}
\usepackage{amsfonts}
\usepackage{bbm}
\usepackage{bm}
\usepackage{hyperref}
\usepackage{arydshln}
\usepackage{multirow}

\title{Interlock-Free Multi-Aspect Rationalization for Text Classification}

\author{Shuangqi Li \and Diego Antognini \and Boi Faltings\\
  École Polytechnique Fédérale de Lausanne, Switzerland\\
  \texttt{firstname.lastname@epfl.ch}}

\begin{document}
\maketitle
\begin{abstract}
 Explanation is important for text classification tasks. One prevalent type of explanation is rationales, which are text snippets of input text that suffice to yield the prediction and are meaningful to humans. A lot of research on rationalization has been based on the selective rationalization framework, which has recently been shown to be problematic due to the interlocking dynamics \citep{yu2021understanding}. In this paper, we show that we address the interlocking problem in the multi-aspect setting, where we aim to generate multiple rationales for multiple outputs. More specifically, we propose a multi-stage training method incorporating an additional self-supervised contrastive loss that helps to generate more semantically diverse rationales. Empirical results on the beer review dataset show that our method improves significantly the rationalization performance.
\end{abstract}

\section{Introduction}
Text classification is a common application of deep neural models \citep{kim-2014-convolutional, conneau-etal-2017-deep}. However, lack of interpretability of the predictions is preventing deep models from being applied in critical fields. A prevalent way of explaining the predictions of text classification is selective rationalization. The key idea is to select informative text snippets of the input texts. If they are short~and coherent enough to be understood by humans and suffice to yield the prediction as a substitute of the full text, they are called rationales \citep{lei-etal-2016-rationalizing}.

\begin{figure}[]
  \centering
  \includegraphics[width=0.48\textwidth]{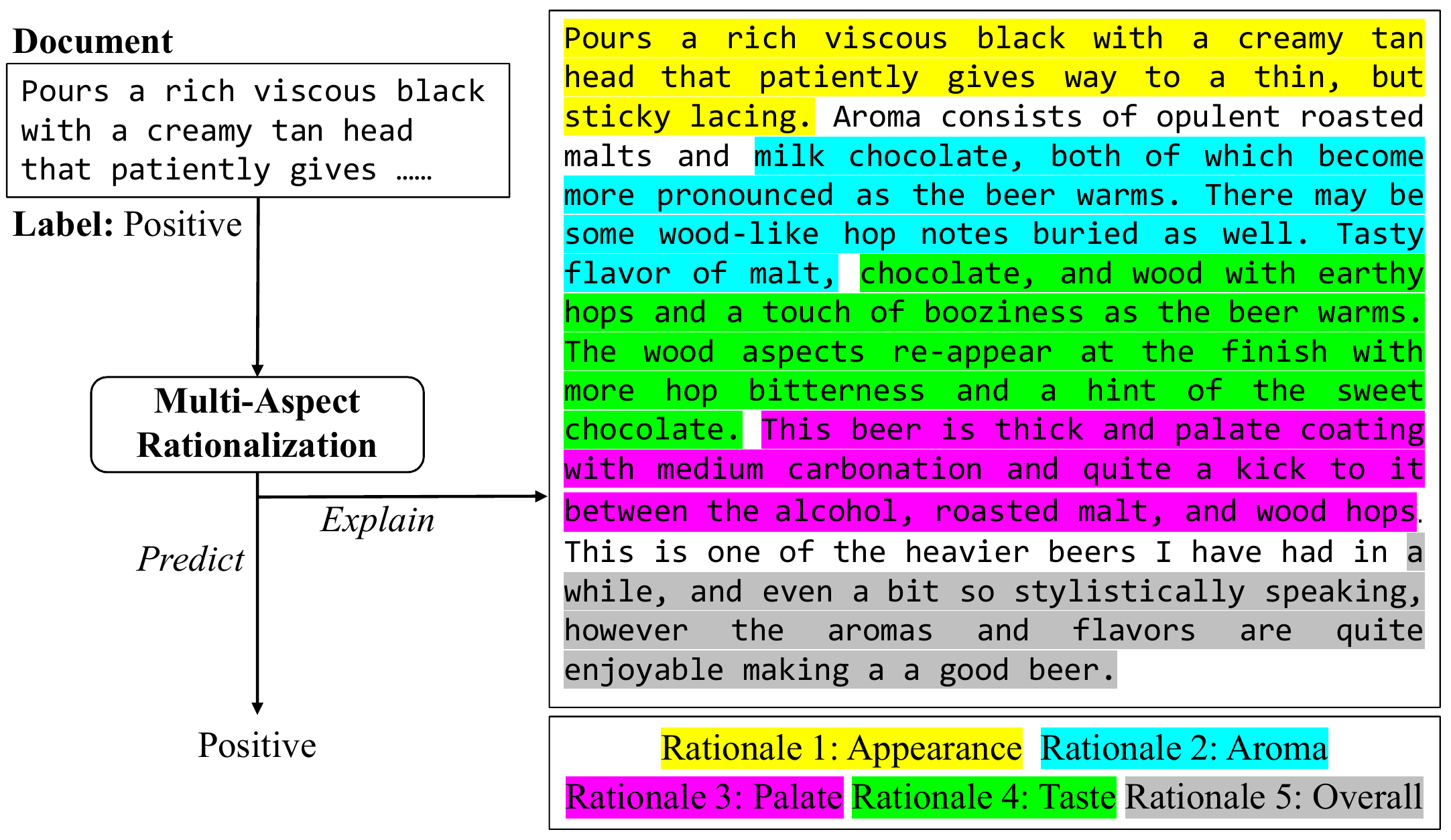}
  \caption{An illustration of multi-aspect rationalization. Given a beer review, the model generates five text snippets (i.e. rationales) that relate to different aspects of the beer, from which the prediction is computed. Different aspects are highlighted in different colors.}
  \label{fig:fig1}
\end{figure}

A line of research has focused on models that are inherently interpretable, i.e, able to produce the prediction along with rationale(s) or mask(s). \citet{lei-etal-2016-rationalizing} proposed the first selective rationalization models that extract one chunk of text as an overall rationale to explain the prediction. Many works \citep{gametheoretic, yu-etal-2019-rethinking,invariantchang,antognini-faltings-2021-rationalization,multidimexpla} follow this framework. Useful explanation can also be multi-aspected \citep{multidimexpla,antognini-faltings-2021-rationalization}, where each aspect is related to a particular concept, as illustrated in Figure \ref{fig:fig1}. Unlike training multiple single-aspect rationale models in order to explain multiple outputs, one can train a single multi-aspect model. A significant advantage is that it only requires the overall label instead of labels for all aspects. This makes multi-aspect rationalization more practical.

Many works rely on the selective rationalization framework \citep{lei-etal-2016-rationalizing} that consists of a generator and a predictor or its variants. Intuitively, the generator extracts a snippet of text from the input and feeds it to the predictor to yield the classification. Training is essentially maximizing the mutual information between the selected text and the label. However, \citet{yu2021understanding} reveal the \textbf{interlocking} problem of this framework: the generator and predictor may get stuck in a suboptimal equilibrium. The interlocking dynamics prevents the generator from selecting the most informative text, and also prevents the predictor from seeing and predicting based on the most informative text.

In this work, we propose a new multi-stage training method that avoids the interlocking problem. The method optimizes different objectives in three stages, incorporating a new self-constrastive loss function, which also promotes more semantically meaningful rationales. Experiments on the beer review dataset show that our multi-stage training fixes the interlocking problem and improves significantly the rationalization performance. Moreover, in a fully unsupervised setting, we show that the generator can learn even better rationalization using only the self-supervised contrastive loss.% (i.e., ignoring the cross-entropy).

%We find that the self-supervised contrastive loss can serve alone as the training objective to achieve good rationalization. With the aid of it, a multi-stage training method that can avoid the interlocking problem is proposed by us. Our experiments on the beer review dataset show that a) even without the label (i.e., in a fully unsupervised setting), the generator can learn almost as good rationalization with the self-supervised contrastive loss, and b) our interlock-free method can achieve a significantly better rationalization.

% In this work, our main contributions are:

\section{Related Work}
\textbf{Selective rationalization} \quad \citep{lei-etal-2016-rationalizing} proposes the generator-predictor framework for rationalization. The generator can select rationales in a soft or hard way. Many works \citep{yu-etal-2019-rethinking,invariantchang,antognini-faltings-2021-rationalization} use a hard constraint, forcing the generator to select text with a pre-specified length. \citet{lei-etal-2016-rationalizing}, \citet{bastings-etal-2019-interpretable} also propose to use a soft constraint to specify the sparsity level instead of the length. \citet{multidimexpla} propose to use a soft probabilistic mask and enable a more flexible rationalization with specified continuity level and sparsity level. The problems in the selective rationalization have raised attention. \citet{invariantchang} show that maximizing the mutual information can be problematic because it may pick up spurious correlations between input features and the output. \citet{10.5555/3454287.3455189} propose a game theoretic approach that captures the multi-faceted nature of rationales. \citet{yu2021understanding} reveal a major problem with the selective rationalization framework that impedes its performance on both classification and rationalization - model interlocking.

\begin{figure}[b]
  \centering
  \includegraphics[width=\columnwidth]{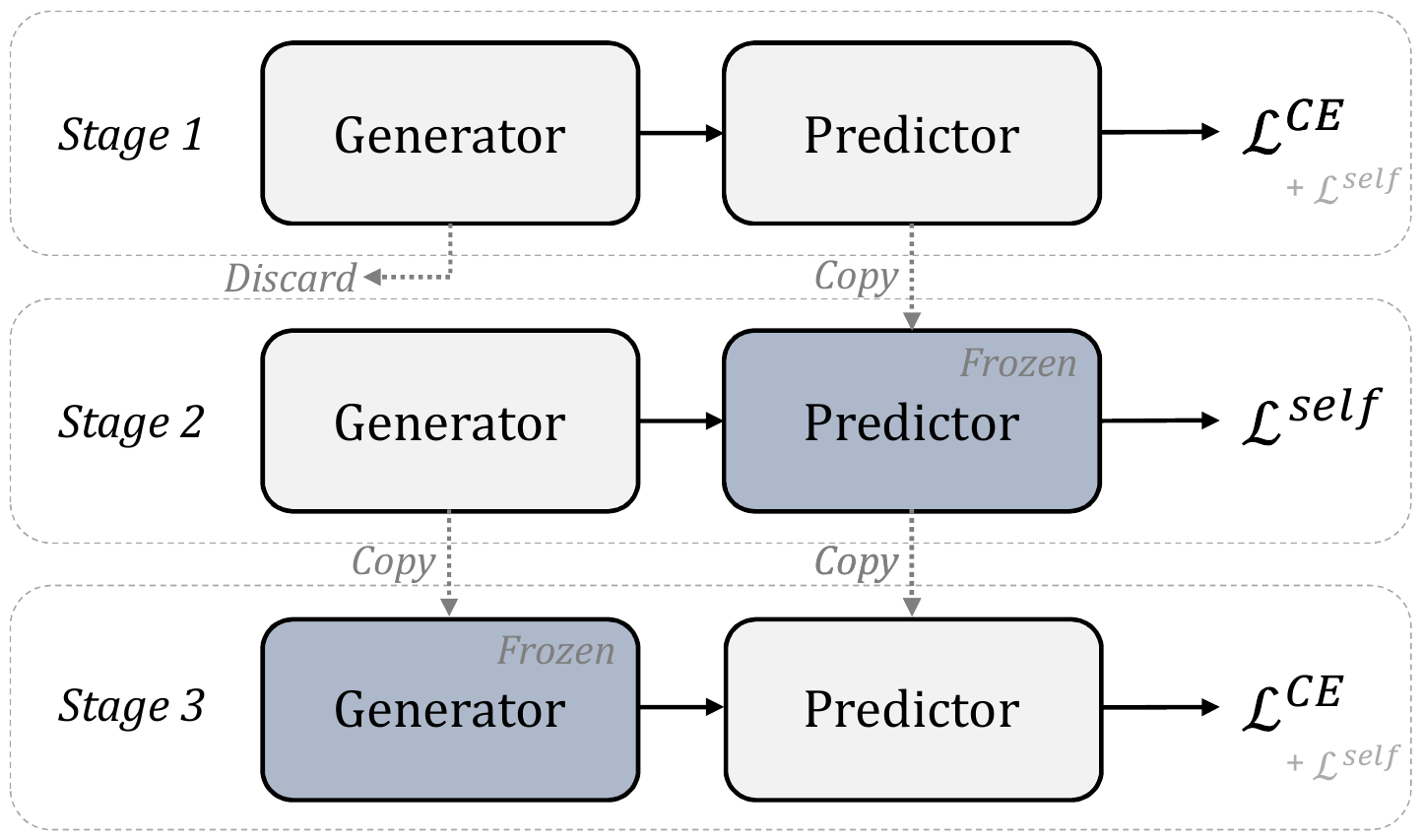}
  \caption{Our proposed interlock-free three-stage training method. Darker modules are frozen.}
  \label{fig:3stage}
\end{figure}

\textbf{Multi-aspect rationalization} \quad \citet{multidimexpla} proposed the first multi-aspect rationalization model, MTM, which uses a probabilistic multi-dimensional mask (one for each aspect) to explain multiple targets (one label for each aspect), and achieves higher classification accuracy and rationalization F1-score at the same time, even in the case of highly correlated data. \citet{antognini-faltings-2021-rationalization} proposed a more practical multi-aspect model, ConRAT, that only requires the overall label. ConRAT identifies a set of concepts (candidate rationales) in the document and builds a selector to decide which ones are chosen, then aggregates the selected concepts to predict a single target variable.

% \textbf{Interlocking} \quad \citet{yu2021understanding} reveals a major problem with the selective rationalization framework that impedes its performance on both classification and rationalization - model interlocking. Ideally, the best generator $ g(\cdot)$ and predictor $ f(\cdot)$ can both be reached at the same time during the training. The predictor $ f(\cdot)$ may overfit to the suboptimal rationales generated by the generator $ g(\cdot)$ and keep reinforcing the generator's suboptimal behaviour. The cause of the interlocking is argued to be that
% $\mathcal L^{CE}(g, f^*)$, where $f^* = \arg\min_{ f} \mathcal L^{CE}(g, f)$,
% is \textbf{concave} with regard to $g$. Intuitively, the predictor can only sees what the generator selects and tends to overfit to the selection. Consequently, the predictor may produce a higher cross entropy loss even when the generator selects a better rationale than the current suboptimal one because the predictor has never seen the better rationale. Then the generator and predictor may get stuck in the suboptimal equilibrium. \citet{yu2021understanding} also proposed a framework called \textit{A2R} that combines the selective rationalization paradigm and the attention-based explanation paradigm, where the concavity in the selective rationalization is mitigated or canceled by the convexity in the attention-based explanation. However, \textit{A2R} cannot guarantee that the interlocking problem is completely avoided and it requires tuning of the hyperparameter that controls the extent of added convexity.

\section{Method}
\subsection{Multi-Aspect Rationalization}
\label{MAR}
%Our multi-aspect rationalization model mostly follows the architectures proposed by \citet{multidimexpla} and \citet{antognini-faltings-2021-rationalization}.
Let $x$ denote the input text, composed of $L$ words $(x^0, x^1, ..., x^{L-1})$. The ground truth is a binary label $y$, telling the overall sentiment (positive or negative). $K$ is predefined as the number of rationales to generate, which is equal to the number of aspects. The architecture is composed of two parts: 1) a rationale \textbf{generator} $ g(\cdot)$ that takes $x$ as input and outputs $K$ rationales; 2) a \textbf{predictor} $ f(\cdot)$ composed by a shared encoder followed by $K$ binary classifiers. The shared encoder produces a representation for each selected rationale per aspect, the classifiers give a prediction for each aspect, which are then linearly aggregated into the final outcome.

% It is worth mentioning that a) our model uses a soft-mask generator following \citet{multidimexpla}, which enables variable length of rationales and their discontinuity to some extent, and hence more flexible rationalization; b) we follow \citet{antognini-faltings-2021-rationalization} and use only the overall label instead of the $K$ labels of all aspects, because the situation without aspect-wise labels is more realistic and common. 

% New version of the above paragraph
To be more specific about the generator, \citet{multidimexpla} provide variable-length and soft rationales but use as many labels as rationales, while \citet{antognini-faltings-2021-rationalization} leverages only the single overall label but is limited to hard rationales with strict continuity and predefined length. However, our model uses a soft generator to enable a more flexible rationalization, and we leverage only the overall label since the situation without aspect-wise labels is more realistic and common. In other words, we take the best of both worlds.

\subsection{The Interlocking Dynamics}
\citet{yu2021understanding} reveal a major problem with the selective rationalization framework - model interlocking. Ideally, the best generator $ g(\cdot)$ and predictor $ f(\cdot)$ can both be reached at the same time during the training. The predictor $ f(\cdot)$ may overfit to the suboptimal rationales generated by the generator $ g(\cdot)$ and keep reinforcing the generator's suboptimal behaviour. More formally, %The cause of the interlocking is argued to be that
$\mathcal L^{CE}(g, f^*), \text{where} f^* = \arg\min_{ f} \mathcal L^{CE}(g, f)$
% \begin{equation}
% \begin{aligned}
% \mathcal L^{CE}(g, f^*), \text{where} f^* = \arg\min_{ f} \mathcal L^{CE}(g, f)
% \end{aligned}
% \end{equation}
is \textbf{concave} with regard to $g$. Intuitively, the predictor can only sees what the generator selects and tends to overfit to the selection. Consequently, the predictor may produce a higher cross entropy loss even when the generator selects a better rationale than the current suboptimal one because the predictor has never seen the better rationale. Then the generator and predictor may get stuck in the suboptimal equilibrium. \citet{yu2021understanding} also proposed a framework called \textit{A2R} that combines the selective rationalization paradigm and the attention-based explanation paradigm, where the concavity in the selective rationalization is mitigated or canceled by the convexity in the attention-based explanation. However, \textit{A2R} is for single-aspect rationalization and cannot guarantee that the interlocking problem is completely avoided. Finally, it requires tuning a parameter to control the extent of added convexity.

\subsection{Self-Supervised Contrastive Loss}
One important feature in multi-aspect rationalization is to have diverse and discriminative rationales. We propose to use contrastive loss in our multi-stage training. Instead of the unsupervised contrastive loss applied in SimCLR \citep{SimCLR} or MoCo \citep{MoCo}, our self-supervised contrastive loss is more similar to the supervised contrastive loss \citep{selfsupervisedcontrast}. The unsupervised contrastive loss contrasts an augmented version of each anchor sample against all other samples, regardless of the unavailable true labels, while the supervised contrastive loss is applied in the fully-supervised setting, leveraging the label information. It contrasts a set of samples from the same class against all other samples from different classes. Formally, the supervised contrastive loss \citep{selfsupervisedcontrast} for a batch of size $N$ is
\begin{equation}
\begin{aligned}
    \mathcal L^{sup} &= -\sum^{N}_{i=1} \frac{1}{N_{y_i}} \sum^N_{j=1} \mathbbm{1}_{i \neq j} \cdot \mathbbm{1}_{y_i = y_j} \cdot \mathcal L^{sup}_{ij} \\
    \mathcal L^{sup}_{ij} &= \log \frac{\exp(z_i\cdot z_j / \tau_c)}{\sum^N_{l=1} \mathbbm{1}_{i \neq l} \cdot\exp(z_i \cdot z_l / \tau_c)}
\end{aligned}
\label{eq:sup}
\end{equation}
where $N_{y_i}$ denotes the number of samples that have the same label as the i\textsuperscript{th} sample in the batch, $z_i$ denotes the representation for the i-th sample in the mini-batch, $\tau_c$ is the temperature hyperparameter.

In our multi-aspect setting, we can consider each rationale per aspect as a sample, and the generator gives $K$ samples (rationales) for each input text. The label of the sample is its aspect index, which is naturally available because we know which rationale is generated for which aspect. This is where the \textit{self-supervision} comes from. The self-supervised contrastive loss is calculated as in Equation~\ref{eq:sup} with the $K \cdot N$ samples in $K$ classes and each class has $N$ samples.
% Do we need a figure to illustrate the loss here?

% Formally, the self-supervised contrastive loss for a mini-batch of size $N$ is
%  \begin{equation}
%  \begin{aligned}
%      \mathcal L^{self} &= \sum^{N}_{i=1} \sum^K_{k=1} \mathcal{L}_{ik}^{self} \\
%      \mathcal L^{self}_{ik} &= -\frac{1}{N}\sum^N_{j=1} \mathbbm{1}_{i \neq j}  \cdot \log \frac{e^{\frac{z_{ik}\cdot z_{jk}}{\tau_c}}}{\sum^N_{l=1} \mathbbm{1}_{i \neq l} \cdot e^{\frac{z_{ik} \cdot z_{lk}}{\tau_c}} + \sum^N_{l=1}\sum^K_{k^{'}=1} \mathbbm{1}_{k \neq k^{'}} \cdot\exp(z_{ik^{'}} \cdot z_{lk^{'}} / \tau_c)},
%  \end{aligned}
%  \end{equation}
%  where $z_{ik}$ denotes the the representation vector for the rationale of the k-th aspect in the i-th sample in the mini-batch. The representation vectors of all aspects come from the output of the shared encoder in the predictor.

\begin{table*}[t!]
\centering
\begin{tabular}{clcccccccc}
\hline
& & \textbf{Avg. Len.} & \textbf{Acc.} & \textbf{Avg. F1} & \textbf{App. F1} & \textbf{Aro. F1} & \textbf{Pal. F1} & \textbf{Tas. F1} & \textbf{Ove. F1} \\
\hline
\multirow{3}{*}{\rotatebox[origin=c]{90}{\textit{\textbf{Long}}}} & Vanilla & 35.5 / 25.1 & 91.7 & 42.4 & 57.7 & 37.0 & \textbf{26.7} & 29.2 & 61.4 \\
& Contra & 35.5 / 25.1 & - & 45.0 & 62.0 & \textbf{43.1} & 20.7 & \textbf{38.7} & 60.7  \\
& 3Stage & 35.5 / 25.1 & 91.7 & \textbf{45.6} & \textbf{63.5} & 41.6 & 26.3 & 26.0 & \textbf{70.9} \\\hdashline[2pt/3pt]
\multirow{3}{*}{\rotatebox[origin=c]{90}{\textit{\textbf{Short}}}} & Vanilla & 23.7 / 18.8 & 90.9 & 34.4 & 37.0 & 31.4 & 21.3 & 33.2 & 26.9 \\
& Contra & 25.4 / 20.6 & - & 36.1 & 58.9 & 41.4 & 20.8 & \textbf{35.1} & 24.3  \\
& 3Stage & 23.9 / 19.9 & 90.6 & \textbf{46.5} & \textbf{59.8} & \textbf{48.1} & \textbf{28.7} & 27.5 & \textbf{68.3} \\\hline
\end{tabular}
\caption{Main results of different training methods on the two modes \textit{Long} and \textit{Short}. The average length is reported on both test set and annotation set. F1 is reported for all aspects described in section \ref{DSet}. \textbf{Bold} marks~best.}
\label{tab:results}
\end{table*}

\subsection{Three-Stage Training}
\label{TST}
Training the model directly with cross entropy or self-supervised contrastive loss suffers from the interlocking problem. We propose a three-stage training framework that alleviates the interlocking problem. 
As illustrated in Figure \ref{fig:3stage}, the model is trained with the cross entropy and self-supervised contrastive loss together in the first stage. In the second stage, the generator is re-initialized, and is trained with the self-supervised contrastive loss $ L^{self}$ with the \textbf{predictor frozen} and only the generator can be updated. In the third stage, the model is trained with the cross entropy and the self-supervised contrastive loss again, with the \textbf{generator frozen} and only the predictor can be updated. 

In the second stage, the objective to be optimized is no longer dominated by the same loss as in the first stage (cross entropy), and only the generator can be updated while the predictor is frozen, therefore they cannot be \textbf{inter}-locked. The generator learns to select rationales that are semantically far apart from each other with the contrastive loss. Intuitively, the shared encoder of the predictor is trained to learn the representation of rationales in the first stage, which are fully exploited in the second stage using a different loss. Similarly, the objective to be optimized in the third stage is not dominated by the contrastive loss as in the second stage , and only the predictor can be updated, therefore the generator and predictor cannot be \textbf{inter}-locked.

One may wonder why the cross entropy dominates in the first stage instead of the contrastive loss. In our situation, we have empirically observed that the cross entropy can be optimized to around the same value with or without the contrastive loss together, but the contrastive loss can be optimized to a significantly lower value alone without the cross entropy. This suggests that the cross entropy dominates the contrastive loss in our settings.

It is worth mentioning that it is essential to optimize different objectives in three stages. In the second stage, for example, if we optimize the same loss, it is still interlocked (or locked) because the objective to be optimized keeps the concavity with regard to the generator. Even worse, it gets less likely to jump out of the suboptimal than before because the predictor is frozen, which means the concave loss landscape for the generator becomes fixed and the generator can get stuck more easily.

\section{Experiments}

\subsection{Datasets}
\label{DSet}
We train and evaluate the classification and rationalization performance on the multi-aspect beer reviews dataset \citep{beerdataset}. Each review describes five aspects related to beer: \emph{appearance}, \emph{aroma}, \emph{palate}, \emph{taste} and \emph{overall}. For each aspect, a rating $ \in \{0.2, 0.3, ..., 0.9, 1.0\}$ is given (but our model uses only the overall rating). Following prior works \citep{bao-etal-2018-deriving}, we binarized the ratings by considering ratings $\leq 0.4$ as negative and ratings $\geq 0.6$ as positive. The number of positive and negative samples are around the same. 60,000 balanced samples are sampled. There are 994 reviews with sentence-level aspect annotations.
\subsection{Training Details}
For all experiments, we used the 200-dimensional GloVe word embeddings \citep{pennington-etal-2014-glove} trained on Wikipedia. We used the Adam optimizer \citep{DBLP:journals/corr/KingmaB14} with a learning rate of 0.0001 and the batch size is 250. The temperature for contrastive loss is $\tau_c = 0.07$. All models are trained with a pre-defined number of epochs and the checkpoint with the minimum loss on validation is evaluated on the test set (for classification accuracy) and annotation set (for rationalization F1-score). Following prior works \citep{antognini-faltings-2021-rationalization}, we manually map the rationales to the aspect ordering that leads to the best F1-score.

\subsection{Results}

Table \ref{tab:results} show results on the beer reviews dataset obtained by evaluating the  model described in Section \ref{MAR} trained by three different methods: \textit{Vanilla}, \textit{Contra}, and \textit{3Stage}. \textit{Vanilla} is the baseline model that adopts the single-stage training with the cross entropy loss, which is the most common loss in previous works \citep{lei-etal-2016-rationalizing, antognini-faltings-2021-rationalization}. \textit{Contra} also adopts the single-stage training but with the self-supervised contrastive loss solely. It is trained without labels, thus not able to do classification. \textit{3Stage} is our proposed three-stage training method described in Section \ref{TST}. All models are evaluated on two modes: \textit{Long} and \textit{Short}, which have different lengths of rationales on average. Specifically, in \textit{Long} mode, the rationales are longer and all tokens are selected in one of the rationales, while in \textit{Short} mode, the rationales are shorter. It is worth mentioning that \textit{Short} models are trained in \textit{Long} mode in the first stage so that a better representation can be learned with a wider horizon for the shared encoder.

Compared with \textit{Vanilla} models, \textit{Contra} can generate (slightly to moderately) better rationales in both \textit{Short} and \textit{Long} modes even without ground-truth information. This suggests that the self-supervised contrastive loss alone is a good objective for learning rationalization, thanks to its ability to promote semantic diversity among rationales. Compared with \textit{Vanilla} models, \textit{3Stage} achieves a moderately higher average F1-score in \textit{Long} mode and significantly higher in \textit{Short} mode, while reaching the same level of classification accuracy. Particularly, the F1-score is significantly higher in terms of the \textit{overall} aspect in both modes. This suggests that \textit{3Stage} can effectively jump out of suboptimal equilibrium and see the more informative text.

% We can observe that the self-supervised contrastive loss alone (\textit{Contra-Long} and \textit{Contra-Short}) is a good objective for the generator to learn to generate rationales, since their rationalization F1-scores are even slightly better than \textit{Vanilla}'s. We can see that \textit{3Stage} achieves a significantly higher average F1-score than \textit{Vanilla} while keeping the same level of accuracy.

\section{Conclusion}
% The interlocking problem impedes the training of selective rationalization models
\vspace{-0.475em}
We proposed a multi-stage training method that avoids the interlocking problem. Its key ingredient is an additional contrastive loss that guides the learning of diverse rationales. In our multi-aspect scenario, experiments on the beer review dataset show that our method achieves significantly better~rationalization. %For future work, we hope that our multi-stage training can generalize to single-aspect rationalization.


\bibliographystyle{acl_natbib}

% \appendix

% \section{Example Appendix}
% \label{sec:appendix}

% This is an appendix.

\end{document}